\documentclass[conference]{IEEEtran}
\usepackage{times}

\usepackage[numbers]{natbib}
\usepackage{multicol}
\usepackage[bookmarks=true]{hyperref}
\usepackage{graphicx}
\usepackage{caption}
\usepackage[bottom]{footmisc}
\usepackage{dblfloatfix}
\usepackage{amsmath}
\usepackage[usenames, dvipsnames]{color}
\usepackage{multirow}
\usepackage[table,xcdraw]{xcolor}
\usepackage{tabularx}
\graphicspath{ {graphics/} }

\usepackage{hyperref}

\newcommand\given[1][]{\:#1\vert\:}

\begin{document}

\title{Grounding Spatio-Semantic Referring \\ Expressions for Human-Robot Interaction}


\author{\authorblockN{Mohit Shridhar}
\authorblockA{School of Computing\\
National University of Singapore\\
Email: mohit@u.nus.edu}
\and
\authorblockN{David Hsu}
\authorblockA{School of Computing\\
National University of Singapore\\
Email: dyhsu@comp.nus.edu.sg}}


%

\twocolumn[{%
\renewcommand\twocolumn[1][]{#1}%
\maketitle
\vspace{-20pt}

\begin{center}
    \centering
    \includegraphics[width=1.0\textwidth]{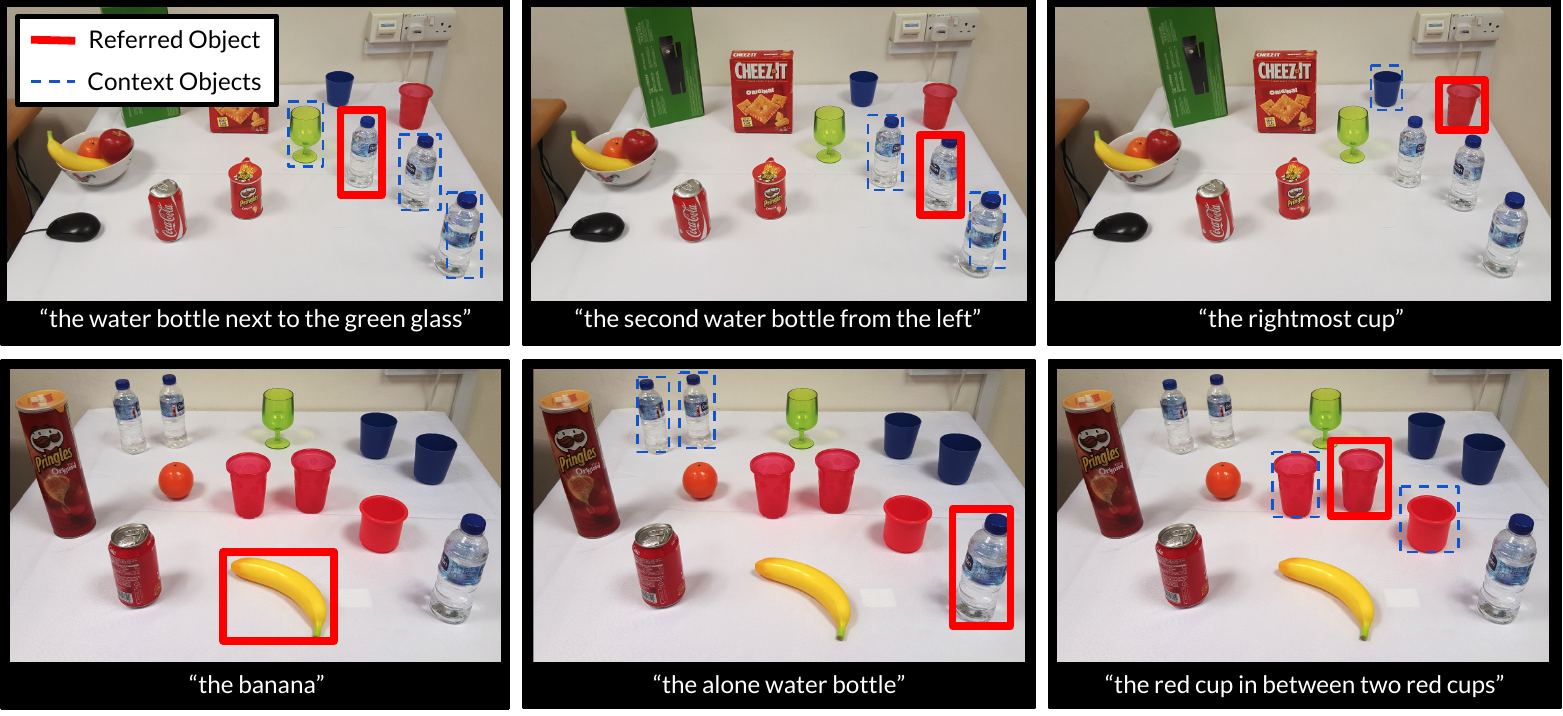}
    \captionof{figure}{Natural language object retrieval in uncluttered environments.}
    
    \label{fig:intro}
    \vspace{10pt}
\end{center}%
}]

\begin{abstract}
The  human language is one of the most natural interfaces for humans to interact with robots. 
This paper presents a robot system that retrieves everyday objects with unconstrained natural language descriptions. A core issue for the system is \textit{semantic and spatial grounding}, which is to infer  objects and their spatial relationships from images and natural language expressions.  
We introduce a two-stage neural-network grounding pipeline that maps natural language referring expressions directly to objects in the images. 
The first stage uses visual descriptions in the referring expressions to generate a candidate set of relevant objects. The second stage examines all pairwise  relationships between the candidates 
and predicts the most likely referred object according to the spatial descriptions in the  referring expressions. A key feature of our system is that by leveraging a large dataset of images labeled with text descriptions, it allows unrestricted object types and  natural language referring expressions.  
Preliminary results indicate that our system outperforms a  state-of-the-art object comprehension system on standard benchmark datasets. We also present a robot system  that follows a user's voice commands to pick and place previously unseen objects.
\end{abstract}

\IEEEpeerreviewmaketitle

\section{Introduction}

In the near future, robots will `live' with humans, providing a variety of services at homes, in workplaces, or on the road.  
To become effective and trustworthy partners,
it is essential for robots to  understand humans' natural language  instructions. 
Doing so reliably requires shared understanding of the environment between humans and robots. A  manifestation of this challenge is the task of \emph{object retrieval}: locating objects in a scene based on natural language expressions.

The key issue with grounding expressions is that language is inherently abstract and free-form in structure. Consider the object retrieval scenario in Fig. \ref{fig:intro}.  Humans can effortlessly generate, interpret and execute an instruction like ``pick up the water bottle next to the green glass''. Any variation of this referring expression e.g: ``the first bottle from the left'' or ``the leftmost blue bottle'' causes minimal confusion to humans. But reliably grounding such free-form object descriptions for robot planning still remains a significant technical challenge. In this scenario, the robot must have two capabilities to locate the referred object: infer semantic information that uniquely describes an object (``water bottle''), and disambiguate multiple instances of the same object using spatial relationships (``on the left''). Given a \emph{spatio-semantic} expression, i.e. a sentence that contains both visual and spatial information e.g: ``the rightmost red cup'', ``the blue bottle in the middle'', the robot must find a specific object in the scene that best fits the description.


Previous works in robotics \cite{wong2015data, gunther2013building, pangercic2012semantic, guadarrama2013grounding, eppner2016lessons} have addressed semantic object retrieval by training classifiers to recognize objects from a set of known categories. But this approach is fallible to variations in the user's description e.g: ``blue water bottle'' vs. ``blue colored bottle'', and typically requires building a custom dataset of possible objects. While attribute decomposition \cite{UW_RSE_ICML2012,fitzgerald2013learning} and probabilistic categorization \cite{sun2016recognising} can ease some of the restrictions of hard classification, retrieval systems are still limited to a small variety of objects. Spatial relationships have been largely studied independently from semantic grounding \cite{paul2016efficient, tellex2010grounding, tellex2012toward} by assuming a pre-existing world model with classified objects and accurate location information. These methods often require an explicitly defined set of possible relationships and constraints e.g: left, right, top, inside, outside etc. Such restrictions limit the capability and robustness of real-world HRI systems. The problem is roughly two-fold: retrieving objects without a pre-specified set of object classes, and disambiguating multiple instances of the same object without pre-defined spatial relationships. 

In this paper, we present a purely image-based approach for grounding unconstrained referring expressions. Recent advancements in deep-learning have paved the way for captioning systems, i.e. end-to-end networks trained on large datasets of image-expression pairs to generate sentence-length descriptions of images \cite{chen2015mind, vinyals2015show, johnson2016densecap}. So instead of constructing a fixed set of classes and spatial templates, we invert the standard captioning model and treat `captions' as object labels, thus rephrasing the retrieval problem as one of caption comprehension. We decompose the retrieval process into two stages: semantic and spatial inference. The first stage uses a captioning model trained on visual descriptions (e.g: ``blue water bottle'') to retrieve all instances of objects mentioned in the query. The second stage uses another captioning model trained on spatio-visual descriptions (e.g: ``left bottle'') to select a single instance based on the specified spatial configuration. The full pipeline takes in a RGB image and a string as input, and predicts a 2D bounding box for the most likely referred object.




We evaluate the accuracy of our retrieval model on RefCOCO's spatio-semantic dataset \cite{kazemzadeh2014referitgame}. And also demonstrate\footnote{Video: \url{https://youtu.be/-qWupV5NhwY}} the pipeline on a simple manipulation task. The implementation includes voice-commanding a 6-DOF MICO arm robot to pick and place unseen items e.g: ``pick up the left blue cup'' and ``put it in the orange cup'', without any explicit object classification or prior spatial knowledge about the scene.


\section{Related Work}

\begin{figure*}[!t]
	\centering
	\includegraphics[width=1.0\textwidth]{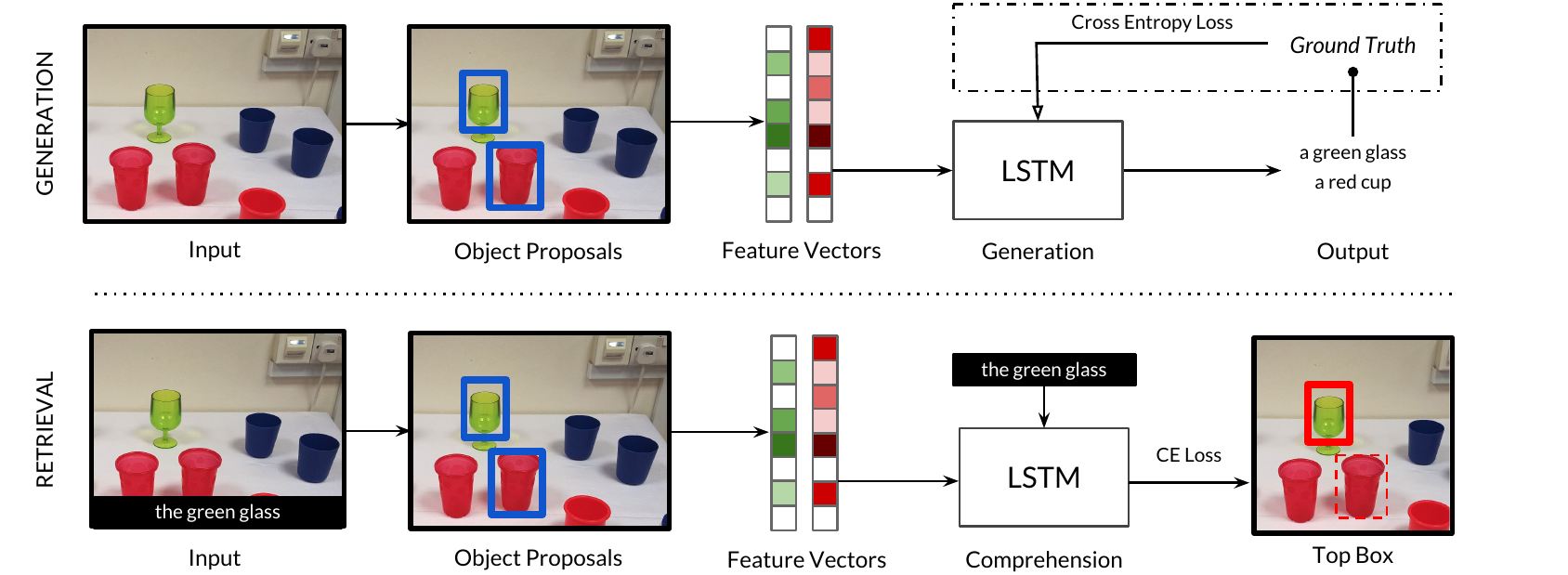}
	\captionsetup{justification=centering,margin=0.0cm}
    \caption{(Top) Caption generation model, dotted box indicates training time. (Bottom) Object retrieval with a pre-trained model. \\ CNN feature extractors and LSTMs share the same weights.}
    \label{fig:gen_comp}
    \vspace{-10pt}
\end{figure*}

\subsection{Semantic Retrieval}

Traditionally in robotics, semantic object retrieval has been treated as a classification problem. This involves using a prior map or database of object classes and super-classes to categorize unseen items  \cite{wong2015data, gunther2013building, blodow2011autonomous, pangercic2012semantic, rottmann2005semantic}. This requires explicitly defining hierarchical relationships (``water bottle'' is a subcategory of ``bottle'') and generic attributes (`blue' colored ``bottle'') or creating unique categories for each object (``blue water bottle'') \cite{guadarrama2013grounding, eppner2016lessons}.  These approaches have several limitations in real-world scenarios. Firstly, the number of categories and attributes are limited in order to reduce misclassifications and search time. While attribute decomposition methods e.g: `is object' and `is blue' and `is cup' \cite{UW_RSE_ICML2012,fitzgerald2013learning} improve the grounding flexibility, they are still incapable of handling rich descriptions e.g: ``cup with a smiley face''. Secondly, users typically have to restrict their descriptions to fit the classifier's input structure e.g: ``blue colored water bottle'' might not equate to ``the blue bottle''. Lastly, the process of encoding visual information usually involves hand-picking feature extractors such as SIFT \cite{lowe2004distinctive} or SURF \cite{bay2006surf}. Our approach to semantic understanding is inspired by recent advances in image caption generation. With the advent of ImageNet \cite{krizhevsky2012imagenet}, deep learning methods have dominated classification \cite{russakovsky2015imagenet, szegedy2015going} and captioning tasks \cite{chen2015mind, vinyals2015show}. The state-of-the-art image captioning systems utilize Convolutional Neural Networks (CNNs) to encode visual features and Recurrent Neural Networks (RNNs) to generate full sentences that describe image regions e.g: ``a blue bottle on the table''. Trained on vast visual datasets of crowd-sourced annotations \cite{krishna2016visual, lin2014microsoft}, these systems automatically extract abstract representations of real-world objects, which encode shapes, colors and other high-level features without any explicit formulation. Further, the same models can also be used for object retrieval, i.e. to find a region on the image that best describes a given sentence \cite{johnson2016densecap, hu2016natural}. Our system exploits this classification-free framework of captioning models to directly map a large corpora of visual descriptions to images. 

\subsection{Spatial Inference}

Spatial inference has also been approached as a classification problem \cite{guadarrama2013grounding, golland2010game, huo2016natural}. With an annotated dataset, it is possible to learn a mapping from sentences to a set of predefined spatial templates e.g: left\_of(reference\_obj, target\_obj). But akin to semantic classification, these methods are too rigid for practical use-cases. The state-of-the-art spatial grounding system \cite{paul2016efficient} utilizes a probabilistic framework instead. This Adaptive Distributed Correspondence Graph (ADCG) model is capable of handling highly abstract expressions with cardinality and ordinality constraints e.g: ``pick up the middle bottle in the row of five bottles on the right''. While these abstractions allow for a high degree of expressivity, as noted by Li et al. \cite{li2016spatial} and Vries et al. \cite{de2016guesswhat} humans predominantly use semantic information (names, shapes, colors etc. that uniquely describe an object) before resorting to complex spatial relationships. In fact, Bisk et al. \cite{bisk2016natural} found that using just spatial relationships without any semantic information dramatically reduces the grounding accuracy. ADCG also requires an explicitly defined set of relations and constraints e.g: left, right, inside, outside etc. Further, generating and interpreting intricate relationships is a difficult task even for humans. Instead Li et al. \cite{li2016spatial} noticed that users prefer an algorithmic approach, i.e. breaking down complex relations into a sequence of simple relations, over a  one-shot method.  Based on these insights, we adopt a much simpler approach towards spatial inference by assuming that most expressions avoid sophisticated relationships. Following Nagaraja et al. \cite{nagaraja2016modeling}, we model spatial relations as binary correspondences e.g: `next to' in ``the water bottle next to the green glass''. This involves training a separate captioning model that takes in multiple image-region features and bounding boxes to handle expressions with non-visual descriptions \cite{nagaraja2016modeling, yu2016modeling, yu2016joint}. Such models are trained on customized datasets: Google Refexp \cite{mao2016generation} and UNC ReferIt  \cite{kazemzadeh2014referitgame}, that contain annotations with both semantic and spatial information e.g: `the left water bottle'', ``the second cup from the left''. Our system uses a spatio-semantic captioning model to directly map expressions with spatial information to image regions, without the need for pre-defined spatial templates or classified objects.




\section{Spatio-Semantic Object Retrieval}

Our objective is to find a bounding box region $R$ on an image $I$, given a sentence $S$. By assuming that $I$ represents the user's perspective of the scene, we can use $R$ for planning robot actions. The challenge is to locate $R$ from $S$ without any pre-defined object classes or spatial templates. For instance, in the expression ``the water bottle next to the green glass''  in Fig. \ref{fig:intro}, we need to find all regions of `water bottle' and `green glass' without training a classifier to explicitly detect a bottle or a glass cup. Subsequently, we also need to disambiguate between multiple water bottles without explicitly defining a spatial template for `next to'. 

In this section, we present our two-stage approach for retrieving objects with spatio-semantic expressions. The first stage locates a set of context objects with a semantic captioning model trained to generate visual descriptions of images. Then we introduce a clustering method to choose the most relevant objects for spatial inference. Finally, the second stage uses a separate captioning model trained to generate spatio-semantic descriptions  to select a single object that best satisfies the specified spatial configuration. We also present an overview of the entire pipeline and discuss potential directions for future work.




\subsection{Caption Generation and Retrieval} \label{semantic_search}

\begin{figure*}[!t]
	\centering
	\includegraphics[width=1.0\textwidth]{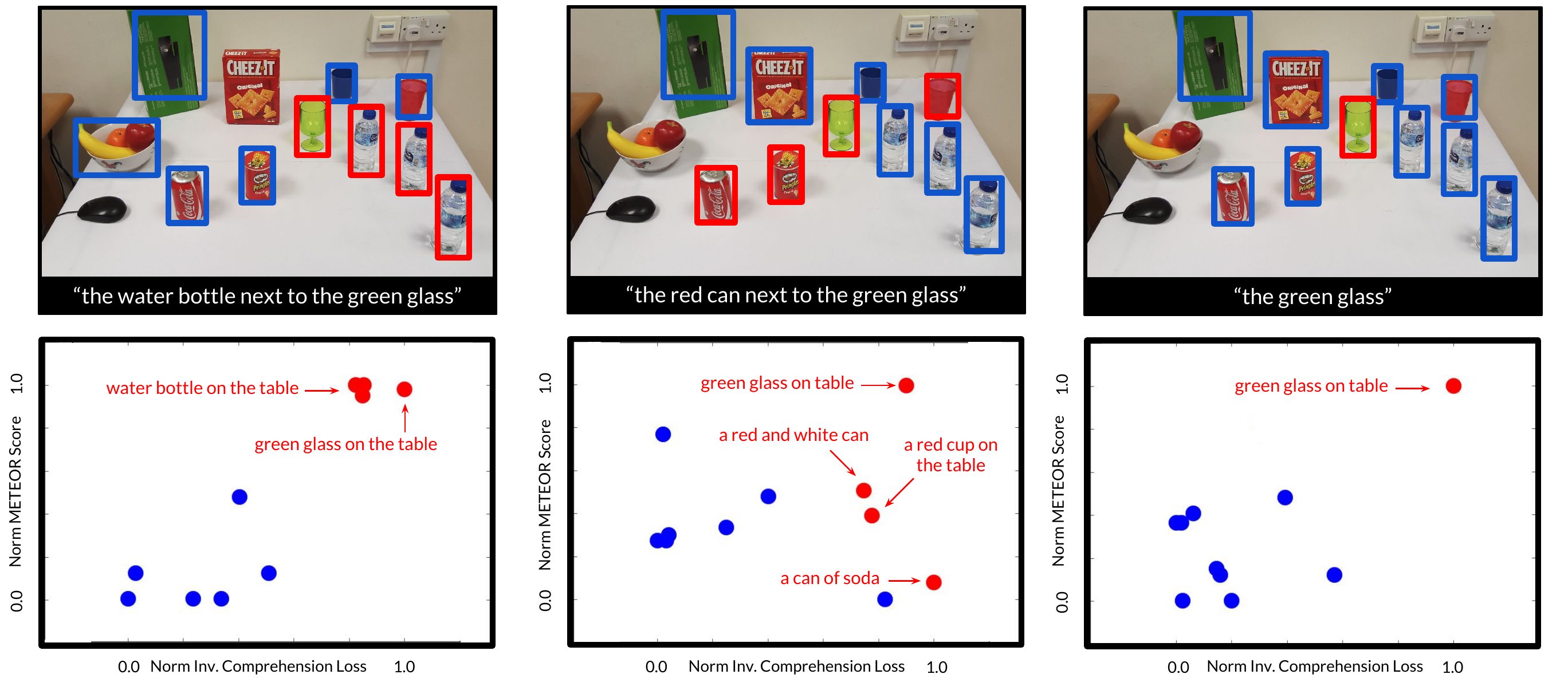}
	\captionsetup{justification=centering,margin=0.6cm}
    \caption{Relevancy Clustering with Top-10 Regions. \\ \textcolor{red}{Red}: relevant objects (point labels indicate generated captions), \textcolor{blue}{Blue}: irrelevant objects.}
    \label{fig:cluster}
    \vspace{-10pt}
\end{figure*}

To avoid explicit classification with a fixed set of categories, we build a generic understanding of a wide variety of objects. By learning abstract representations of common physical characteristics, we can potentially infer details about unseen objects. The state-of-the-art techniques in generic visual understanding use vast amounts of human-annotated datasets to generate captions for images. We exploit this direct mapping between descriptions and images to retrieve relevant objects.

The underlying architectures for caption generation and object retrieval are closely related. In fact, a model trained for captioning can also be used for retrieval. We use a pre-trained model provided by Densecap \cite{johnson2016densecap}, which was built on the Visual Genome dataset \cite{krishna2016visual}. The dataset consists of $\sim$100k images with an average of 43.5 region annotations per image e.g: ``cats play with toys hanging from a perch'' and ``woman pouring wine
into a glass''. Our choice of this captioning model is motivated by the large size and the detailed expressivity of the annotations used for training. Further, the dataset is composed of objects from $\sim$80k categories, making the model applicable to a diverse range of real-world scenarios. Densecap's model extracts CNN features using a standard VGG-16 network \cite{simonyan2014very} from a set of regions proposals generated by a Faster R-CNN \cite{ren2015faster} style architecture. The output of the network's conv5 layer is fine-tuned to encode each region as a 4096-length feature vector. These feature vectors are fed into a LSTM to generate a new word at each time-step as an one-hot encoded vector. During training, the LSTM weights are recomputed based on the cross-entropy loss between the generated encoding sequence and the ground truth one-hot encoding from the dataset (see Fig. \ref{fig:gen_comp}). The model has vocabulary of 10,497 words with a vector embedding size of 512 for each word. The whole network is trained end-to-end on a NVIDIA Titan X GPU for $\sim$ 3 days.

Formally, generation can be represented as $ {\mathrm{argmax}}_{S} \; p(S \given[\big] R, I) $, where $S$ denotes a sentence, and $R$ describes a region on the image $I$. Comprehension, or object retrieval can then be represented as finding a region $R^{*} = \mathrm{argmax}_{R \in C}\; p (R \given[\big] S, I)$ where $R$ is sampled from a set of regions from context $C = \{R_{1}, R_{2} ... R_{n} \}$. By applying Bayes rule and assuming a uniform prior for $p(R \given[\big] I)$, Mao et al. \cite{mao2016generation} show that:

\begin{equation} \label{eq:equality}
  \underset{R \in C}{\mathrm{argmax}}\; p (R \given[\big] S, I) = \underset{R \in C}{\mathrm{argmax}}\; p (S \given[\big] R, I) \\
\end{equation}

\noindent
The intuition behind Eq. \ref{eq:equality} is that both $p (S \given[\big] R, I)$ and $p (R \given[\big] S, I)$ are dependent on the same feature vectors; the abstract encoding that uniquely represents an object is the same regardless of generation or comprehension. And the LSTM simply acts as a translator between the feature encoding and the raw sentence. Following this insight, the comprehension model closely resembles the generation model except for the final step (see Fig. \ref{fig:gen_comp}). Instead of computing the cross-entropy loss $L$ between the generated sequence and the ground truth, we compute the loss between the generated sequence and the user-provided query. The region that results in the lowest loss is chosen as the most likely referred object:  

\begin{equation} \label{eq:loss}
\underset{R \in C}{\mathrm{argmax}}\; p (R \given[\big] S, I) = \underset{R \in C}{\mathrm{argmin}}\; L (R, S, I)
\end{equation}
While Eq. \ref{eq:loss} picks a single region as the most relevant object, expressions may refer to more than just one object or contain multiple instances of the same object. So instead, we select top $k$ regions with the lowest loss scores, assuming each region contains a single object with minimal overlap. Based on our observations, $k=10$ is sufficient for most scenarios as expressions rarely refer to more than 10 objects. With this method, we can extract most of the relevant context objects in a given scenario without any explicit classification, e.g: from ``the water bottle next to the green glass'' we can retrieve all instances of ``water bottle''s and ``green glass''es.

\begin{figure*}[!t]
  \centering
  \includegraphics[width=1.0\textwidth]{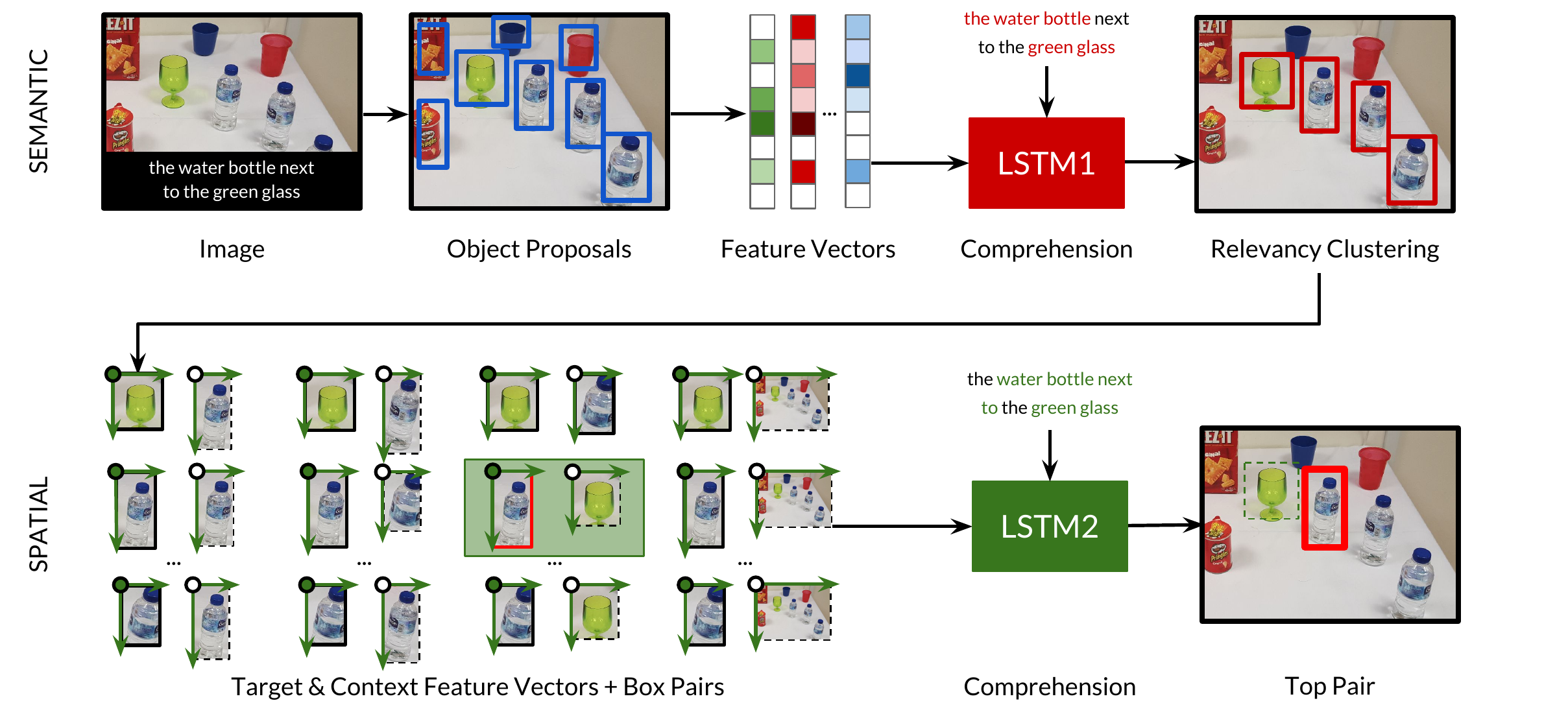}
  \captionsetup{justification=centering,margin=0.6cm}
    \caption{Complete pipeline from image and query to final bounding box.}
    \label{fig:pipeline}
    \vspace{-10pt}
\end{figure*}

\subsection{Relevancy Clustering} \label{clust}

Even in uncluttered environments, most of the objects in the scene are irrelevant to the user's query. The semantic retrieval module returns top $k$ object matches for a given visual description. But it is difficult to predict the number of referred objects in a query. Fixing $k=10$ is often sufficient for retrieving all relevant objects. But most expressions will likely refer to less than 10 objects. As shown in Fig. \ref{fig:cluster}, most of the top-10 regions are irrelevant to the query. Later in Section \ref{pipeline} we note that a large number of irrelevant objects will increase the compute time and reduce the accuracy of spatial inference. Choosing a smaller $k$ will decrease the number of irrelevant objects, however we risk disregarding some relevant objects. Hence there is a need for dynamically choosing $k$ depending on the scene. 


We propose a novel clustering method to address this issue. By virtue of the object retrieval method, the loss scores of relevant objects will be much lower than that of irrelevant objects. So it is possible to segment out the two groups assuming there is discernible loss difference in-between them. We also want to pick objects, which the model is most confident about. But we cannot directly use the magnitude of the loss score as a confidence metric as it is susceptible to variations in lighting, image size and other conditions. So in conjunction with the comprehension loss, we compute a generative distance score between the model generated caption and user-provided caption using METEOR \cite{banerjee2005meteor}. METEOR is standard machine translation metric, which calculates a normalized distance score between two sentences e.g: ``the green glass'' and ``the green cup'': 0.90, ``the green glass'' and ``the blue book'': 0.20, by comparing synonyms and paraphrase matches. This metric allows us to estimate the generative confidence, i.e. how similar the model generated caption of the object is to the actual user description. We use K-means clustering \cite{kanungo2002efficient} with $K=2$ to segmented out relevant objects using these two metrics (see Fig \ref{fig:cluster}). Both scoring metrics are normalized, and the comprehension loss is inverted such that relevant cluster is closer to $\{1,1\}$. The resulting objects in the top cluster are then passed to the spatial grounding module.

\subsection{Spatial Grounding} \label{spatial}

Natural expressions that refer to a singular object can be sufficiently located using the semantic retrieval model presented in Section \ref{semantic_search}. However, in environments with multiple instances of the referred object, spatial cues are necessary to disambiguate between similar instances. This requires examining relationships between neighboring objects. Following the insights from Nagaraja et al. \cite{nagaraja2016modeling} and Bisk et al. \cite{bisk2016natural}, we assume these relationships are predominantly binary in nature. This is reasonable since a lot of simple spatio-semantic expressions follow a target-context structure e.g: ``water bottle next to the green glass'' context: ``green glass'', target: ``water bottle''. Although this binary model fails to infer cardinality, ordinality constraints e.g: ``the third water bottle from the left'' and more sophisticated relationships, the assumption is sufficient for a large number of use-cases. Even seemingly non-binary associations such as ``the leftmost bottle'' in a scene with more than two bottles, can be modeled as a binary relation with respect to the entire image.  

Spatial inference also requires considering perspective constraints \cite{li2016spatial}. Human generated expressions can take various perspectives: object-centric (``bottle next to the glass''), ego-centric (``bottle on my left''), robot-centric (``bottle on your right'') or image-centric (``the bottle on the top left of the image''). In this work, we only address object-centric and image-centric perspectives. For real robot experiments, we assume the user's perspective is roughly aligned with the camera so that we can treat ego-centric as image-centric perspectives. We leave other perspectives for future work.

Our spatial grounding model closely follows UMD RefExp's \cite{nagaraja2016modeling} approach. Formally, we extend our objective from finding a single region $R^{*}$ to finding a pair $(R_{t}, R_{c})$ where $R_{t}$ is the target region and $R_{c}$ is the context region:

\begin{equation} \label{eq:pair}
R^{*}_{t} = \underset{R_{t} \in C}{\mathrm{argmax}}  \bigg\{ \underset{R_{c} \in C \setminus R_{t} }{\mathrm{max}}\;  p (R_{t}, R_{c} \given[\big] S, I) \bigg\}
\end{equation}
\noindent
We estimate $p (R_{t}, R_{c} \given[\big] S, I)$ with a separate LSTM, which takes in pairs of feature vectors $f_{t}, f_{c}$ and bounding boxes $b_{t}, b_{c}$  (all extracted during semantic retrieval). The bounding boxes are encoded in a normalized format $b = [\frac{x_{min}}{W}, \frac{y_{min}}{H}, \frac{x_{max}}{W}, \frac{y_{max}}{H}, \frac{Area_{box}}{Area_{image}} ]$, where $W$ and $H$ are the image width and height respectively. All features and boxes are concatenated into a single vector $[f_{t}, b_{t}, f_{c}, b_{c}]$ and fed into the LSTM. Each target region $\{f_{t}, b_{t}\}$ is paired with $k-1$ context regions $\{f_{c1}, b_{c1}, f_{c2}, b_{c2}, ... f_{c(k-1)}, b_{c(k-1)}\}$ from the clustering output. The last context region $f_{cI}, b_{cI}$ is the whole image. For each object, the first $k-1$ context regions address local object-centric perspectives (e.g: ``to the left of''), while the whole image addresses global image-centric perspectives (e.g: ``the leftmost''). In total, there are $k^{2}$ pair-wise comparisons. So in implementation, we set the upper-bound of $k$ as 10 to limit the rapid growth in compute time.  Furthermore, instead of directly picking the top target-context pair from Eq. \ref{eq:pair}, we integrate information from $k-1$ pairs using a noisy-or function \cite{nagaraja2016modeling}:

\begin{equation} \label{eq:noisyor}
R^{*}_{t} = \underset{R_{t} \in C}{\mathrm{argmax}} \bigg\{ 1 - \prod_{R_{c} \in C \setminus R_{t}} (1 - p (R_{t}, R_{c} \given[\big] S, I)) \bigg\}   
\end{equation}
\noindent
Although we primarily model binary relationships, both the whole image context and the noisy-or function help approximate some non-binary relations. Future works could address the problem of directly inferring extra-binary relationships.

During training we use the negative-margin model proposed by Nagaraja et al. \cite{nagaraja2016modeling} to penalize negative target-context pairs. These negative pairs are constructed by randomly sampling pairs with a incorrect $R_{t}$ from the dataset. The model was trained on UNC's RefCOCO dataset \cite{yu2016modeling}, which consists of $\sim$140k referring expressions for 50k objects in 19k images. On average the dataset contains 3.9 same-type objects per image, thus forcing annotations to include more than just appearance-based descriptions e.g: ``the second man on the left''. Some expressions include size-based comparisons e.g: ``the larger bottle'' and others include 3D spatial cues e.g: ``the bottle behind the glass'' despite the annotations being purely 2D bounding boxes. The model has vocabulary of 2,020 words with a vector embedding size of 1024 for each word. Training takes $\sim$ 1 day on a Titan X GPU.

The object proposals and feature vectors extracted during semantic retrieval are reused during spatial inference. During training, the spatial module uses ground-truth bounding boxes from the dataset and feature vectors extracted from Densecap \cite{johnson2016densecap}. The 4096-length feature vectors are fine-tuned to 1000-dimensional vectors before feeding them into the LSTM. The $R^{*}_{t}$ from the top noisy-or pair is chosen as the final referred region.

\subsection{Pipeline} \label{pipeline}
Fig. \ref{fig:pipeline} shows the complete pipeline that integrates all modules: semantic search, relevancy clustering and spatial inference. As noted in the previous sections, the two LSTMs specialize in two different tasks. LSTM1 is trained to find all instances of the objects mentioned in the query using pure visual information. It ignores any spatial cues provided in the description as the training dataset \cite{krishna2016visual} mostly contains appearance-based annotations. LSTM2 specializes in both semantic and spatial inference, but with a strong emphasis on spatial relationships. As shown by Bisk et al. \cite{bisk2016natural}, pure spatial inference without semantic information dramatically reduces the grounding accuracy, so LSTM2 needs to jointly infer both aspects. However, since LSTM1 was trained on a much larger dataset with highly detailed visual descriptions, its semantic inference capability is superior to that of LSTM2. Naively feeding in all pairs of objects into the spatial module decreases the grounding accuracy as LSTM2 might get confused with unfamiliar objects due to inferior semantic knowledge. And so LSTM1 is used as a precursor to choose relevant objects for LSTM2. It is possible to combine both the datasets and just train the second stage of the pipeline in an end-to-end manner. However, the datasets have to be carefully balanced so that the larger semantic dataset doesn't overshadow the spatial inference during learning. We leave this for future works. 

\section{Results}

In this section, we evaluate the accuracy of our model on validation datasets for spatio-semantic expressions. We also include implementation details and qualitative results from our experiments with a MICO arm-robot.

\subsection{Evaluation}

\newcolumntype{P}[1]{>{\centering\arraybackslash}p{#1}}
\setlength\extrarowheight{3.2pt}

\begin{table}[t]
\centering 
\normalsize
\begin{tabular}{|p{2.1cm}|c|c|c|c|}
\hline
\multicolumn{5}{|c|}{\textbf{Comprehension Prec@1 Scores (\%)}}                                                                                                                                                                                         \\ \hline
                         & \multicolumn{2}{c|}{GT Proposals}                                                                          & \multicolumn{2}{c|}{MCG Proposals}                                                                         \\ \cline{2-5} 
\multirow{-2}{*}{Method} & \begin{tabular}[c]{@{}c@{}}UMD\\ RefExp\end{tabular} & \begin{tabular}[c]{@{}c@{}}Our\\ Model\end{tabular} & \begin{tabular}[c]{@{}c@{}}UMD\\ RefExp\end{tabular} & \begin{tabular}[c]{@{}c@{}}Our\\ Model\end{tabular} \\ \hline
\multicolumn{5}{|c|}{\cellcolor[HTML]{C0C0C0}RefCOCO - Val}                                                                                                                                                                                        \\ \hline
Noisy-Or                 & 74.9                                                 & \textbf{76.5}                                       & 55.8                                                 & \textbf{57.1}                                                 \\ \hline
Max                      & 73.7                                                 & \textbf{75.2}                                       & 55.2                                                 & \textbf{56.9}                                                 \\ \hline
\multicolumn{5}{|c|}{\cellcolor[HTML]{C0C0C0}RefCOCO - Test A}                                                                                                                                                                                     \\ \hline
Noisy-Or                 & 74.4                                                 & \textbf{77.0}                                       & 57.4                                                 & \textbf{60.4}                                                 \\ \hline
Max                      & 72.4                                                 & \textbf{75.4}                                       & 56.8                                                 & \textbf{59.4}                                                 \\ \hline
\multicolumn{5}{|c|}{\cellcolor[HTML]{C0C0C0}RefCOCO - Test B}                                                                                                                                                                                     \\ \hline
Noisy-Or                 & 75.6                                        & \textbf{75.7}                                                & \textbf{54.8}                                                 & 54.3                                                 \\ \hline
Max                      & 74.6                                                 & \textbf{75.4}                                       & \textbf{53.9}                                                 & 53.7                                                 \\ \hline
\end{tabular}
\captionsetup{justification=centering,margin=0.1cm}
\caption{Preliminary results for comprehension. Ground-Truth vs. MCG proposals \cite{arbelaez2014multiscale}. Baseline: UMD RefExp \cite{nagaraja2016modeling}. Noisy-Or refers to Eq. \ref{eq:noisyor}, Max refers to Eq. \ref{eq:pair}.}
\label{results}
\end{table}

We evaluate our pipeline on UNC's RefCOCO validation dataset and benchmark the performance against UMD RefExp \cite{nagaraja2016modeling}. The evaluation measures the accuracy at which the model can locate an object region (bounding box on image) with its given spatio-semantic description. The precision scores are computed based on the overlap between the model-chosen box and the ground-truth box provided by the dataset. Boxes with Intersection-over-Union (IoU) $>0.5$ are considered as correct groundings. We use UMD RefExp as our baseline since it produces near state-of-art results for grounding image-based spatio-semantic expressions and also inspired most of our spatial inference module (see Section \ref{spatial}). There are two key differences between the baseline and our approach. Firstly, UMD RefExp uses feature vectors from a pre-trained VGG-16 classifier built for categorizing objects, whereas our model uses fine-tuned feature vectors from a captioning model (Densecap) built for generating visual descriptions. Secondly, for images with more than 10 proposals (i.e. $k > 10$), UMD RefExp randomly picks 9 pairs for spatial inference. However, our model uses relevancy clustering (described in Section \ref{clust}) to sort out a small subset of regions for pair-wise comparison. 

The majority of images in the RefCOCO dataset are of people, but for HRI we are mostly concerned with objects. So we test our model on three separate partitions provided by the dataset: Test A - person-oriented expressions (e.g: ``the guy on the left''), Test B - object-oriented expressions (e.g: ``cup on the right''), and Val - all objects. Since our system is specifically designed for spatio-semantic grounding, we prune expressions with pure spatial information e.g: ``the top left'', ``bottom corner'' by using a POS tagger \cite{toutanova2003feature} to remove sentences without nouns. Test A contains 750 images with 5531 expressions. Test B contains 750 images with 4954 expressions. And Val contains 1500 images with 10557 expressions. Following UMD RefExp, we present results for both ground-truth proposals (human annotated boxes) and generated MCG proposals \cite{arbelaez2014multiscale} to study the accuracy with and without false-positive object proposals\footnote{Even though for real robot experiments we use end-to-end trained proposals generated by Densecap, we use MCG during evaluation for benchmarking purposes}.

Table \ref{results} shows preliminary results for the comprehension task. In majority of the cases, our model outperforms the UMD RefExp baseline. However, based on our observations, most of the improvements seem to be due to the fine-tuned features vectors from Densecap. The relevancy clustering had minimal impact on improving the grounding accuracy. Clustering is only effective in highly cluttered environments where most of the objects are irrelevant to the user's query. And since the validation set contains on average 10.2 ground-truth proposals and 7.4 MCG proposals per image, UMD RefExp's brute-force inference method can most often check all possible pair combinations without reaching the $k>10$ limit. We plan on extending our work to support multi-image retrieval, for which relevancy clustering might be better suited. 

In general, the noisy-or method (Eq. \ref{eq:noisyor}) performed slightly better than the max  method (Eq. \ref{eq:pair}). So despite integrating information from pair-wise comparisons, the relationships learnt were predominantly binary in nature. The person-oriented expressions had the greatest improvement in accuracy, whereas our model had minimal or negative impact on object-oriented expressions. This is probably due to the large bias in the number of person-oriented images in both the datasets. Future works could build a more balanced dataset with a larger variety of common objects. 

Recently, Yu et al. \cite{yu2016joint} proposed a joint speaker-listener-reinforcer model and claim to produce state-of-the-art results on the RefCOCO dataset. Future works could integrate and benchmark our pipeline against this framework when their pre-trained model becomes available.

\subsection{Manipulator Experiments}

To demonstrate a sample use-case of spatio-semantic inference, we integrate our grounding pipeline with a 6-DOF manipulator (Kinova MICO v1). The objective is to pick and place unseen objects purely through voice commands e.g: ``pick up the leftmost blue cup'' and ``put it in the orange cup at the back''. The robot has no prior semantic or spatial information about the scene, and has to infer the user's intent upon receiving a new request. We define two actions: `pick up' and `put it', which have to be stated at the start of each sentence. The rest of the instruction is fed directly into the grounding pipeline. The voice commands are synthesized into text via Amazon Echo's Alexa API. 

Our object-retrieval model takes in a RGB image and a query string as input, and outputs a bounding box to indicate the most likely referred object. However, a 2D bounding box is often insufficient for planning a 6-DOF grasp action. So we use a Kinect RGB-D camera to estimate the 3D location of the object contained inside the box. Using a RGB aligned point-cloud, we estimate the object's centroid by taking an average over all the points\footnote{Note that the 3D position estimate is only used for grasp planning, while the spatial inference module uses a plain RGB image}. We then plan a naive forward grasp or top-down grasp based on the size of the object. Forward grasp involves approaching the object from the back and closing the gripper when the finger joints are in-line with the centroid. Similarly, top-down grasp involves approaching the object from the top with the gripper facing down. If the object is too wide or too short (with respect to the size of the gripper), the planner executes a top-down grasp, otherwise a forward grasp. For placing objects, the end-effector is positioned directly above the referred object and then the gripper is opened. This naive approach was sufficient for most of our pick and place experiments. However, future works could integrate state-of-the-art techniques in manipulating novel objects \cite{levine2016learning, lenz2015deep}. The entire pipeline was executed on a NVIDIA Titan X GPU. In an uncluttered scene with 10-20 objects: the CNN feature extraction takes $\sim$ 1.3 secs, and the query grounding takes $\sim$ 0.15 secs. For both aspects, the compute time grows linearly with the number of objects. The overall voice-to-action delay time is $\sim$ 2-5 secs, which includes retrieving the synthesized text from Amazon's servers, computing the centroids of segmented point clouds, and planning a cartesian path for the 6-DOF arm.

We conducted an informal user-study to assess the fidelity of our HRI system. 8 participants were individually instructed to pick and place a specific item in a scene with 3-15 objects. The only restriction was ``to unambiguously describe the object with the \emph{simplest} expression possible''. The scenes contained a mixture of single-instance and multi-instance objects to provoke the use of both visual and spatial references. Responses from preliminary experiments were highly positive. Users were able to accurately communicate their object of intent even in semi-cluttered environments. The system was also able to locate objects never before seen in the training datasets e.g: ``the orange cup'' by implicitly combining prior visual knowledge of `cups' and `orange' objects. Fig. \ref{fig:intro} shows a representative sample of some successful groundings. Since captioning models are inherently designed to generate descriptions in an open-world setting, the system is capable of comprehending a much larger variety of objects (e.g: dogs, planes, buildings etc.) than the examples presented in this paper. However, due to the physical limitations of the arm robot, we restrict our experiments to a small subset of household objects (e.g: cups, cans, bottles etc.).

Occasionally, the model produced incorrect groundings (see Appendix Fig. \ref{fig:failure} for failure cases). So before grasping an object, the robot asked the users if it had found the right object by pointing at it. If not, it subsequently iterated through the rest of the objects in the order of likelihoods provided by Eq. \ref{eq:noisyor}. Despite initially incorrect groundings, the users found the specified item within 2-3 attempts as the relevancy clustering was effective in narrowing down possible objects of interest. In the future, this simple yes-no correction method could be improved with a proper dialog-driven system \cite{de2016guesswhat} where the robot asks specific questions pertaining to the object. In general, the grounding pipeline was most accurate when the referring expression contained less than 4 types of objects (not including multiple instances) and exactly $1$ or no spatial relationship. Intricate expressions that required inferring multiple relationships e.g: ``the cup right of the green glass and left of the blue bottle'' failed consistently. Although, in many scenarios users could have used simpler binary relationships. Furthermore, during the study, some users requested for more sophisticated actions than `pick up' and `put it'. Similar to the abstraction capability of our grounding pipeline, users suggested defining high-level actions e.g: ``stack'', ``pour'', ``put it inside'' etc, which follow a series of pick and place actions. We plan on conducting a more through user-study with quantitative analysis to evaluate the accuracy of our grounding pipeline in unconstrained real-world scenarios.


\section{Conclusion} \label{sec:conclusion}

In this paper, we present a technique for locating objects using natural language. We use an image-based captioning model to find relevant objects in the scene. Then use a separate captioning model to pick a pair of relevant objects, which best satisfies the specified spatial configuration. Our model outperforms UMD RefExp in grounding spatio-semantic expressions on the RefCOCO dataset. We also demonstrate a simple object manipulation use-case by voice-commanding an arm robot to pick and place objects. Even though we are far from achieving an ideal shared world-understanding between humans and robots, we hope our work takes a significant step in tackling this challenge.

\section*{Acknowledgments}

This work was supported by the NUS School of Computing Strategic Initiatives. We thank members of the Adaptive Computing Lab at NUS for thoughtful discussions. We also thank the reviewers for their insightful feedback.

\bibliographystyle{abbrvnat}
\bibliography{references}

\twocolumn[{%
\renewcommand\twocolumn[1][]{#1}%
\appendix

\vspace{20pt}

\subsection{Failure Cases}
\label{sec:failure}
\vspace{10pt}
\begin{center}
    \centering
    \includegraphics[width=1.0\textwidth]{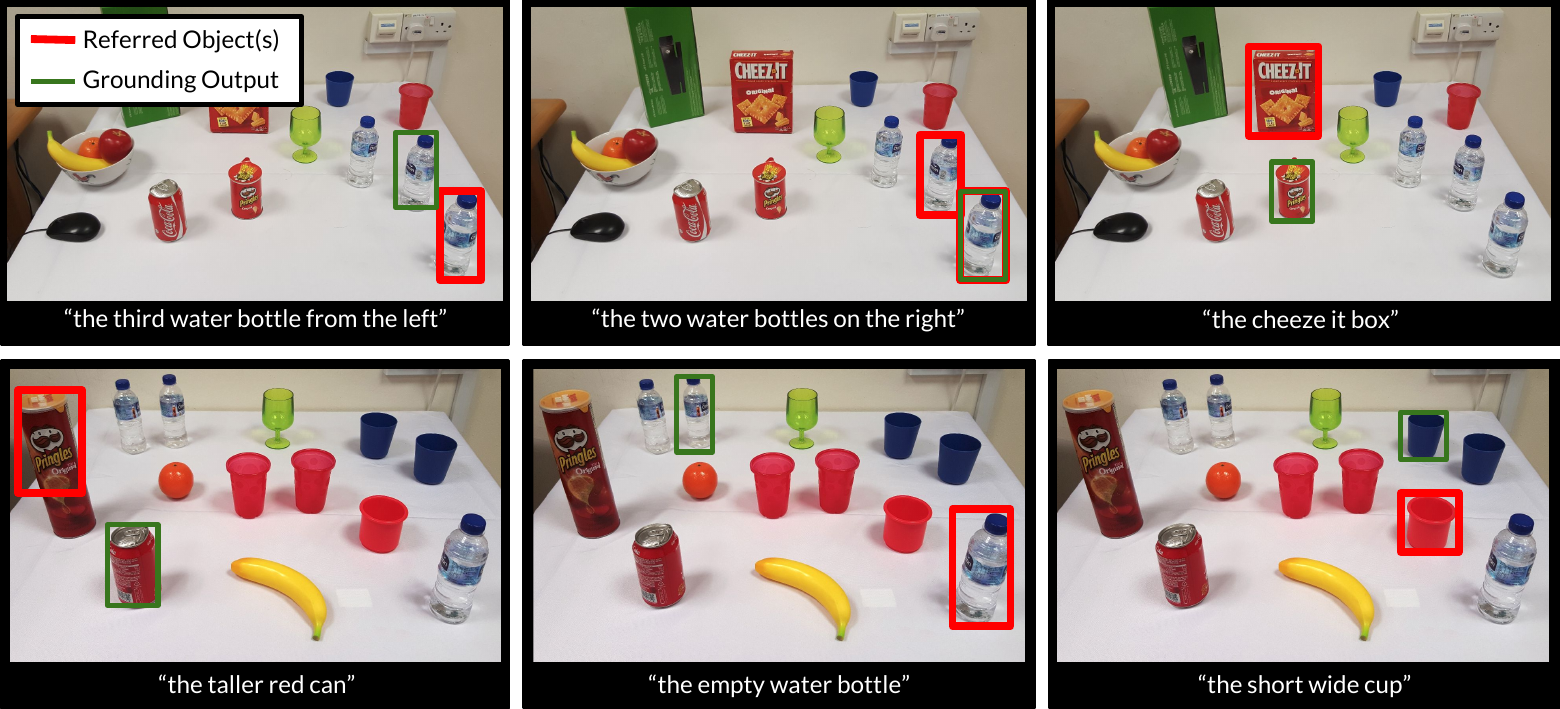}
    \captionsetup{justification=centering,margin=0.05cm}
    \captionof{figure}{Retrieval failure cases. The system fails to infer cardinality (e.g: `two bottles') and ordinality (e.g: `second bottle') constraints. Expressions referring to labels, logos or printed text on objects often produce incorrect groundings. Size based comparisons (e.g: `bigger', `wider', `taller' etc.) are limited by the accuracy of the object proposals.}
    
    \label{fig:failure}
    \vspace{10pt}
\end{center}%

\vspace{5pt}
\subsection{Manipulator Experiment Setup}

\begin{center}
    \centering
    \includegraphics[width=1.0\textwidth]{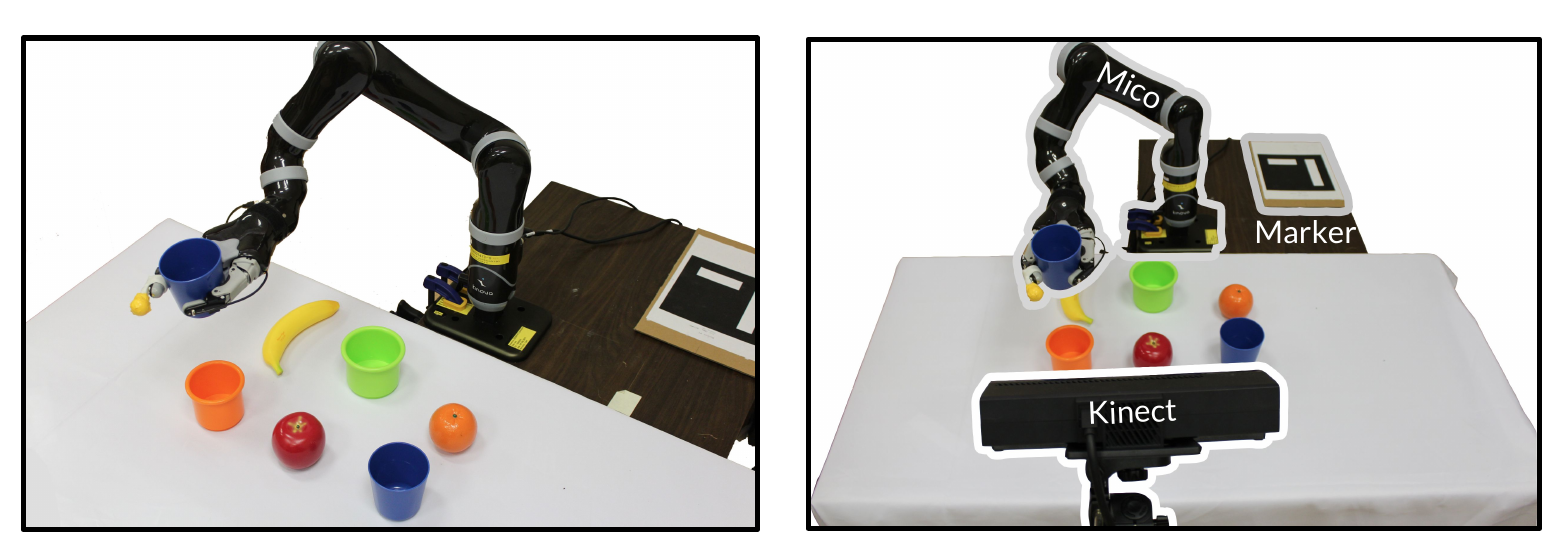}
    \captionsetup{justification=centering,margin=0.05cm}
    \captionof{figure}{Manipulator setup used for experiments. The RGB image from the Kinect V2 camera (right) was used for spatio-semantic grounding. The depth image was used to estimate the object centroids for grasp planning. And the fiducial marker was used to compute the extrinsic offset between the camera and the robot. The users controlling the arm stood behind the Kinect so that their perspectives were roughly aligned with that of the camera.}
    
    \label{fig:mani_setup}
    \vspace{10pt}
\end{center}%

}]
\end{document}